\documentclass[times,twocolumn,final]{elsarticle}

\usepackage{ipcai_arxiv}
\usepackage{framed,multirow}
\usepackage{amssymb,latexsym}
\usepackage{url}
\usepackage{natbib}
\usepackage{amsmath,amssymb,amsfonts,bm}
\usepackage{booktabs}
\usepackage{arydshln}
\usepackage{tabulary}
\usepackage{graphicx,epstopdf}
\usepackage{textcomp}
\usepackage{subcaption}
\usepackage[svgnames,table]{xcolor}
\usepackage[colorlinks=true,citecolor=cyan,urlcolor=cyan]{hyperref}
\usepackage{placeins}
\usepackage{fancyhdr}
\usepackage{float}
\usepackage{makecell}
\usepackage[capitalise]{cleveref}

\graphicspath{{./}}

\newcommand{\win}[1]{\cellcolor{green!15}\textbf{+#1}}
\newcommand{\loss}[1]{\cellcolor{red!10}{$-$#1}}

\fancypagestyle{firstpagestyle}{
    \fancyhf{}
    
    \fancyhead[CO]{\em \fontsize{9pt}{8pt}\selectfont}
}

\begin{document}

\begin{frontmatter}

\title{Artificial Intelligence for Workflow Analysis in Colorectal Surgery: A Multicentric, Cross-Procedural Development and Generalization Study }

\author[1,2,3]{Pietro \snm{Mascagni} \corref{cor}\fnref{co-first}}
\author[1,4]{Julia \snm{Alekseenko}\fnref{co-first}}
\author[1,4]{Pooja P \snm{Jain}\fnref{co-first}}
\author[5]{Marta \snm{Goglia}}
\author[6]{Andrea \snm{Balla}}
\author[7]{Ludovica \snm{Baldari}}
\author[5]{Gianfranco \snm{Silecchia}}
\author[2,3]{Claudio \snm{Fiorillo}}
\author[3,8]{Vincenzo \snm{Tondolo}}
\author[6]{Salvador \snm{Morales-Conde}}
\author[7]{Luigi \snm{Boni}}
\author[2,3]{Sergio \snm{Alfieri}}
\author[1,4]{Nicolas \snm{Padoy}}

\cortext[cor]{Corresponding Author: 
Pietro Mascagni, MD, PhD, Universit\`a Cattolica del Sacro Cuore, Largo Francesco Vito, 1, 00168 Roma RM, Italy. Telephone: +39 3922736975. Email: pietro.mascagni@ihu-strasbourg.eu. ORCID: 0000-0001-7288-3023}
\fntext[co-first]{These authors contributed equally to this work and share first authorship}

\address[1]{IHU Strasbourg, Strasbourg, France}
\address[2]{Surgical Video Analysis, Fondazione Policlinico Universitario Agostino Gemelli IRCCS, Rome, Italy}
\address[3]{Universit\`a Cattolica del Sacro Cuore, Rome, Italy}
\address[4]{University of Strasbourg, CNRS, INSERM, ICube, UMR7357, France}
\address[5]{Department of Medical-Surgical Sciences and Translational Medicine, Faculty of Medicine and Psychology, Sapienza University of Rome, Italy}
\address[6]{University Hospital Virgen Macarena, University of Sevilla, Seville, Spain}
\address[7]{Department of General \& Minimally Invasive Surgery, Fondazione IRCCS - Ca' Granda - Ospedale Maggiore Policlinico di Milano, Milan, Italy}
\address[8]{UOC Chirurgia Digestiva e Colon-Retto, Ospedale Isola Tiberina--Gemelli Isola, Rome, Italy}

\begin{abstract}
\textbf{Background:} Minimally invasive colorectal surgeries (MIS-CRS) are characterised by significant variability and inconsistent outcomes. ColoWorkflow, a tool for the video-based assessment (VBA) of MIS-CRS workflow, was recently validated. However, manual VBA is time-consuming, limiting implementation. This study presents AI-ColoWorkflow, a deep learning model for automated surgical workflow analysis across MIS-CRS.

\textbf{Method:} Operative videos of MIS-CRS were collected from 4 centres and a publicly available dataset. Phases and steps were manually annotated according to ColoWorkflow. A deep learning model combining a fine-tuned DINOv3 vision transformer for per-frame visual feature extraction with a hierarchical multi-stage temporal convolutional network was jointly optimized for phase and step recognition. The model trained on pooled multicentric data, namely AI-ColoWorkflow was compared against centre-specific and procedure-specific models on a held-out test set. The following metrics were used for evaluation: macro F1 score, balanced accuracy, precision, and recall. 

\textbf{Results:} AI-ColoWorkflow achieved a macro F1 of 73.01\%\,$\pm$\,10.27 (balanced accuracy 73.43\%) for phase recognition and 39.82\%\,$\pm$\,7.06 (balanced accuracy 38.65\%) for step recognition. The global model outperformed centre- and procedure-specific models in most experiments except procedure-specific step recognition. In the generalization analysis, mean F1 was 48.42\% for phase recognition. 

\textbf{Conclusion:} AI-ColoWorkflow can reliably recognize MIS-CRS phases. A single model trained on pooled, multicentric, multi-procedural data generalises at least as well as and often better than centre- or procedure-specific models for phase recognition in MIS-CRS, while procedure-specific step models retain advantages for certain procedure types, motivating hybrid training strategies for future surgical AI development. 

\end{abstract}

\end{frontmatter}
\thispagestyle{firstpagestyle}

\bigskip

\section{Introduction}
Minimally invasive colorectal surgery (MIS-CRS) is the gold standard approach for several diseases of the colon and rectum. However, MIS-CRS requires advanced technical skills and is characterised by significant variability in surgical workflows, the sequence of phases and steps composing a procedure, across centres and surgeons \cite{hanna2022lapco}. This variability is associated with inconsistent outcomes, underscoring the need for tools to objectively study and standardise surgical practice \cite{curtis2020association,curtis2021nearmiss}. 

Video-based assessment (VBA) has emerged as a promising approach to analyse and improve surgical procedures by extracting data-driven insights from operative videos \cite{gruter2023videobased}. Recently, ColoWorkflow, a generalizable framework for workflow analysis across minimally invasive colorectal procedures, was developed through a modified Delphi process involving over 40 international experts in colorectal surgery and VBA \cite{jain2026video}. ColoWorkflow defines 9 procedure-agnostic phases and 34 procedure-specific steps and was validated on a multicentre dataset of laparoscopic and robotic colorectal procedures, demonstrating broad applicability and moderate inter-rater agreement  \cite{jain2026video}. However, manual annotation of colorectal surgical workflows remains time-consuming, requiring approximately 60\% of the video duration, which limits scalability and practical implementation.

Artificial intelligence (AI) and computer vision (CV) have been trained to recognise surgical workflows in relatively standardised procedures such as laparoscopic cholecystectomy and gastric bypass \cite{ramesh2021multitask,yuan2024hecvl}. More recently, AI models have also been applied to well-defined procedures within colorectal surgery, including sigmoid resection and rectal surgery \cite{kolbinger2023context,nakajima2024automated}. Yet, no AI model has been developed to perform workflow analysis across different colorectal procedures using a unified, consensus-based ontology.

It was hypothesised that an AI model could learn to apply the ColoWorkflow, recognising phases and steps across minimally invasive colorectal procedures, from right hemicolectomies to rectal resections. This study aimed to develop AI-ColoWorkflow, a deep learning model for automated workflow analysis in MIS-CRS, and to evaluate its ability to generalise across procedures and centres. 

\section{Method}
\begin{figure*}[t]
  \centering
  \includegraphics[width=0.90\textwidth, trim={5mm 55mm 5mm 60mm}, clip]{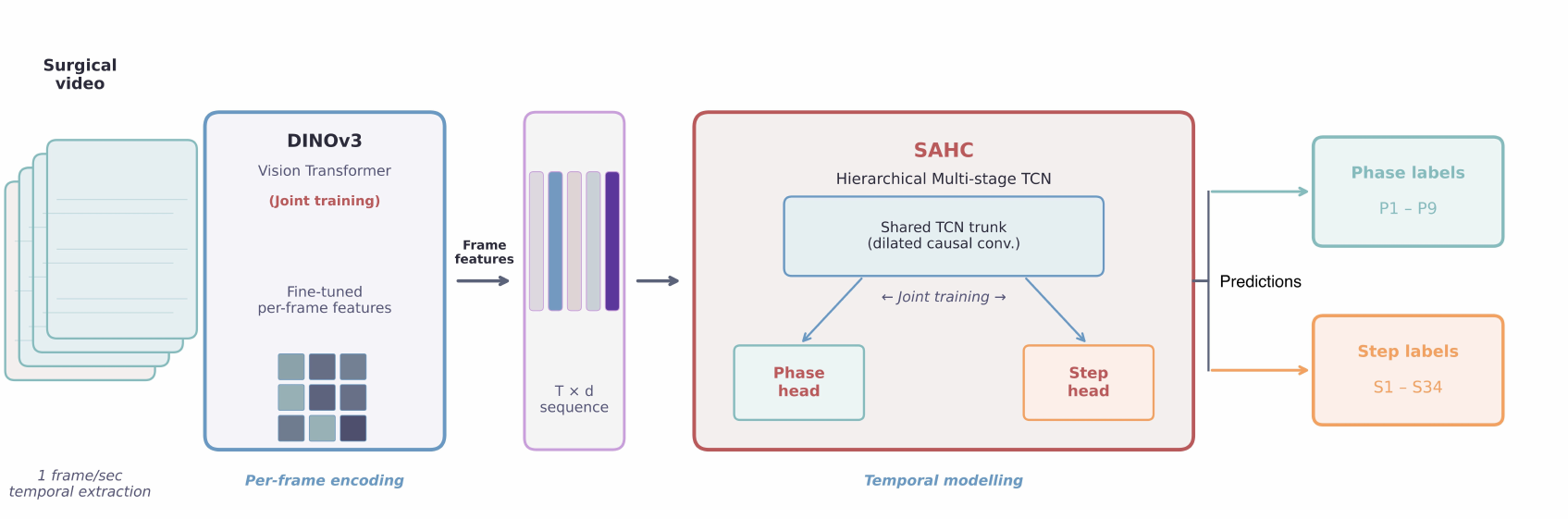}
  \caption{AI-ColoWorkflow system. Two sequential stages: (1) spatial feature extraction with a fine-tuned DINOv3 vision transformer; (2) temporal modelling with the SAHC hierarchical multi-stage TCN, jointly trained for phase and step recognition. d: feature dimension; SAHC: Segment-Attentive Hierarchical Consistency network; T: time; TCN: Temporal Convolutional Network}
  \label{fig:architecture}
\end{figure*}

\subsection{Study design}
This is a development study using multicentric surgical video data collected under ethical approval from Fondazione Policlinico Universitario Agostino Gemelli IRCCS, Rome, Italy (ID 6456). The study is reported in accordance with a version of the QUAIDE (Quality Assessment of pre-clinical AI studies in Diagnostic Endoscopy) modified to account for the non-diagnostic nature of the AI task \cite{antonelli2025quaide}. 

\subsection{Data collection and annotation}
Operative videos of major laparoscopic and robotic MIS-CRS types performed on adult patients at partner centres across Italy, Spain, and France were collected. Videos were de-identified and pre-processed locally at each participating centre using Endoshare, a surgeon-friendly and publicly available desktop application \cite{arboit2026endoshare}. Additional MIS-CRS videos were retrieved from the publicly available HeiCo dataset from Heidelberg, Germany \cite{maierhein2021heico}. De-identified videos were then uploaded to the MOSaiC platform \cite{mazellier2023mosaic}, where a medical doctor trained on VBA performed temporal workflow annotation following the ColoWorkflow framework; annotation guidelines and best practices are detailed in ColoWorkflow Supplementary Material \cite{jain2026video}. 

\subsection{AI-ColoWorkflow}
AI-ColoWorkflow is a deep learning system that maps raw endoscopic video frames to a sequence of surgical workflow labels (phases and steps), as defined by ColoWorkflow \cite{jain2026video}. The pipeline comprises two sequential components: spatial feature extraction and temporal modelling (Figure~ \ref{fig:architecture}). 

For the spatial feature extraction, video frames were extracted at one frame per second and resized to a standard resolution. A DINOv3 vision transformer \cite{simeoni2025dinov3} was fine-tuned on the annotated dataset to extract per-frame visual representations. DINOv3 was selected for its self-supervised pre-training on large and diverse image corpora, which provides strong initialisation for downstream fine-tuning in novel visual domains. Unlike convolutional backbones, the self-attention mechanism of the vision transformer captures global context within each frame, relevant for surgical scenes in which anatomical relationships across the field of view disambiguate phase and step identity. 

For the temporal modelling, frame-level features were passed to the Segment-Attentive Hierarchical Consistency Segment-(SAHC) network \cite{ding2022exploring}, a multi-stage temporal convolutional network using exponentially dilated causal convolutions to capture long-range temporal dependencies without relying on future frames. Each stage refines the predictions of the previous stage through a residual correction mechanism, progressively reducing over-segmentation errors. SAHC is jointly trained with two task-specific output heads, phase and step, sharing a common temporal trunk. A combined cross-entropy loss is optimised over both heads simultaneously, with equal task weighting. This multi-task design explicitly exploits the hierarchical relationship between phases and steps: phase context regularises step predictions, and step activations provide fine-grained signal that improves phase boundary detection. 

\subsection{Experimental setup and statistical analysis}
\label{sec:experimental}

Three main experimental configurations were defined. 

First, a global model, namely AI-ColoWorkflow, was trained on pooled data from all centres and procedure types and evaluated on a held-out test set. This configuration reflects the most optimistic scenario; a single model trained on data from every target institution.  

Second, multi-centre mono-procedure models were trained and evaluated for each individual procedure type (e.g. sigmoid resection only, right hemicolectomy only, etc.) and mono-centre multi-procedures models were developed for each individual centre (e.g., centre 1 only, centre 2 only, etc.). This is an evolution of the classical approach in Surgical Data Science (SDS), where models are trained on single-centre data to perform a task on a given procedure type. Comparing these procedure/centre specific models against the global AI-ColoWorkflow model reveals whether a single diverse training set outperforms smaller, homogeneous datasets, a question with direct implications for how future training data should be collected and organised. 

Third, a leave-one-centre-out (LOCO) cross-validation evaluated whether AI-ColoWorkflow generalises by iteratively withholding one centre from training and evaluating the model on the unseen centre. This design specifically studies generalization, quantifying the performance gap when the model encounters a previously unseen institution with potentially different surgical styles, equipment, and patient populations. 

The dataset was partitioned at the case level into training ($\approx$60\%), validation (20\%), and test (20\%) sets, preventing temporal data leakage. The validation set was used exclusively for hyperparameter tuning; the test set was held out during all optimisations.  

Model performance was evaluated using metrics selected to address the specific challenges of workflow recognition. The macro F1 score, the harmonic mean of precision and recall averaged across all classes, served as the primary outcome metric, as it penalises models that perform well on frequent classes but fail on rare ones. Balanced accuracy was used as it accounts for class imbalance, a critical consideration in surgical procedures where certain phases (e.g. dissection) occupy a disproportionately large share of the video while others (e.g. leak testing) may be very brief. Precision (positive predictive value) quantified the reliability of the model's detections, measuring the proportion of correct predictions among all instances assigned to a given class. Recall (sensitivity) quantified the model's completeness, measuring the proportion of correctly identified instances among all true occurrences.  

Both descriptive and inferential analyses were performed. Results are reported as mean \,$\pm$\ standard deviation across videos. Differences between AI-ColoWorkflow and a majority-class baseline were assessed with the Wilcoxon signed-rank test on per-video macro F1 scores across all 11 test videos
($p<0.05$) was considered significant. The model was implemented in PyTorch \cite{paszke2019pytorch}, with OpenCV \cite{bradski2000opencv} for frame processing and scikit-learn \cite{pedregosa2011scikit} for metrics; hardware specifics and training configuration are further detailed in Table \ref{tab:supp_specs}.  

\section{Results}

\subsection{Dataset}

\begin{table}[t]
\centering
\caption{Dataset splits by centre and procedure type. LH: Left hemicolectomy; RH: Right hemicolectomy; RR: Rectal resection; SR: Sigmoid resection; TPC: Total proctocolectomy.  }
\label{tab:dataset_splits}
\small
\setlength{\tabcolsep}{5.5pt}
\begin{tabular}{@{}llcccc@{}}
\toprule
\textbf{Centre} & \textbf{Surgery type} & \textbf{Train} & \textbf{Val} & \textbf{Test} & \textbf{Subtotal}\\\midrule
Centre 1      & LH & 2 & 1 & 2 & 5 \\
           & RH & 1 & 0 & 0 & 1 \\\cmidrule(lr){2-6}
           & \textit{Sub} & \textbf{3} & \textbf{1} & \textbf{2} & \textbf{6}\\\midrule
Centre 2       & LH & 1 & 1 & 0 & 2 \\
           & RH & 3 & 1 & 1 & 5 \\
           & RR & 1 & 0 & 0 & 1 \\
           & SR & 1 & 0 & 0 & 1 \\\cmidrule(lr){2-6}
           & \textit{Sub} & \textbf{6} & \textbf{2} & \textbf{1} & \textbf{9}\\\midrule
Centre 3    & LH & 2 & 0 & 0 & 2 \\
           & RH & 4 & 1 & 2 & 7 \\
           & RR & 1 & 0 & 0 & 1 \\
           & SR & 0 & 1 & 0 & 1 \\\cmidrule(lr){2-6}
           & \textit{Sub} & \textbf{7} & \textbf{2} & \textbf{2} & \textbf{11}\\\midrule
Centre 4 & SR & 4 & 1 & 1 & 6 \\\cmidrule(lr){2-6}
           & \textit{Sub} & \textbf{4} & \textbf{1} & \textbf{1} & \textbf{6}\\\midrule
HeiCo      & RR & 4 & 2 & 2 & 8 \\
           & SR & 2 & 1 & 1 & 4 \\
           & TPC& 6 & 2 & 2 & 10\\\cmidrule(lr){2-6}
           & \textit{Sub} & \textbf{12}& \textbf{5} & \textbf{5} & \textbf{22}\\\midrule
\rowcolor{gray!10}\textbf{TOTAL}&&\textbf{32}&\textbf{11}&\textbf{11}&\textbf{54}\\\bottomrule
\end{tabular}
\end{table}

\begin{table*}[p]
\centering
\caption{AI-ColoWorkflow performance across phases and steps. All values shown as global\,$\pm$\,std, where global is computed on concatenated test frames and $\pm$ is the standard deviation of per-video scores across individual test videos. P7 (Preplanned additional procedures) and S25 (Ileoanal pouch preparation - intraabdominal part) were not evaluated since these were not represented in the test set.}
\label{tab:unified_results}
\footnotesize
\begin{tabular}{lccc}
\toprule
\textbf{Surgical workflow item} & \textbf{Prec.\ (\%)} & \textbf{Rec.\ (\%)} & \textbf{F1 (\%)}\\\midrule
\rowcolor{gray!10}\multicolumn{4}{l}{\textbf{Phase recognition}}\vspace{1pt}\\
P1. Port placement and abdomen exploration                          & 76.96\,$\pm$\,16.02 & 69.36\,$\pm$\,15.10 & 72.96\,$\pm$\,10.68\\
P2. Vascular dissection and ligation, mesocolon/mesorectum dissection, additional lymphadenectomy         & 80.82\,$\pm$\,13.39 & 72.77\,$\pm$\,16.48 & 76.58\,$\pm$\,12.87\\
P3. Colon and/or rectum mobilisation                                       & 69.83\,$\pm$\,9.62  & 75.55\,$\pm$\,16.64 & 72.58\,$\pm$\,10.64\\
P4. Colorectal transection                                          & 64.07\,$\pm$\,30.09 & 64.16\,$\pm$\,20.42 & 64.11\,$\pm$\,19.07\\
P5. Anastomosis                                                     & 69.50\,$\pm$\,27.02 & 82.08\,$\pm$\,27.81 & 75.27\,$\pm$\,23.85\\
P6. Completion of operation                                         & 82.55\,$\pm$\,23.17 & 55.61\,$\pm$\,31.31 & 66.45\,$\pm$\,27.65\\
P8. Unplanned procedures                                            & 46.49\,$\pm$\,28.75 & 37.44\,$\pm$\,29.56 & 41.48\,$\pm$\,26.89\\
P9. Extracorporeal procedures                                       & 92.04\,$\pm$\,10.17 & 97.59\,$\pm$\,4.82  & 94.73\,$\pm$\,7.35\\\cmidrule{1-4}
\rowcolor{gray!5}\textbf{Phase macro average} & \textbf{73.19\,$\pm$\,11.16} & \textbf{73.43\,$\pm$\,9.49} & \textbf{73.01\,$\pm$\,10.27}\\\midrule
\rowcolor{gray!10}\multicolumn{4}{l}{\textbf{Step recognition}}\vspace{1pt}\\
S1.  Insertion of ports and instruments (Intraabdominal part)                             & 73.42\,$\pm$\,22.24 & 56.22\,$\pm$\,24.17 & 63.68\,$\pm$\,22.15\\
S2.  Abdominal cavity and structures assessment                     & 27.06\,$\pm$\,14.62 & 60.43\,$\pm$\,21.56 & 37.38\,$\pm$\,15.80\\
S3.  Adhesiolysis/structure division for exposure                   & 50.34\,$\pm$\,35.11 & 17.04\,$\pm$\,10.74 & 25.46\,$\pm$\,14.82\\
S4.  Mesentery/mesocolon exposure                                   & 38.14\,$\pm$\,33.31 & 43.29\,$\pm$\,37.30 & 40.55\,$\pm$\,30.00\\
S5.  Peritoneum incision and Toldt’s/Gerota fascia dissection –– left& 37.14\,$\pm$\,32.19 & 41.49\,$\pm$\,31.99 & 39.19\,$\pm$\,28.45\\
S6.  Peritoneum incision and Toldts/Gerota and Fredets  fascia dissection -- right              & 41.77\,$\pm$\,36.55 & 25.07\,$\pm$\,30.15 & 31.33\,$\pm$\,30.19\\
S7.  Inferior mesenteric artery isolation, ligation and division -- left                   & 25.73\,$\pm$\,35.19 & 19.79\,$\pm$\,19.57 & 22.37\,$\pm$\,23.13\\
S8.  Inferior mesenteric vein isolation, ligation and division -- left                   & 33.33\,$\pm$\,0.00  &  7.48\,$\pm$\,0.00  & 12.22\,$\pm$\,0.00\\
S9.  Ileocolic vessels and/or branches isolation, ligation and division -- right    & 41.60\,$\pm$\,1.10  &  9.34\,$\pm$\,0.27  & 15.25\,$\pm$\,0.44\\
S10. Middle colic vessels branches and/or Henle's trunk branches isolation, ligation and division -- right              & 32.85\,$\pm$\,21.55 & 20.96\,$\pm$\,20.80 & 25.59\,$\pm$\,21.17\\
S11. Mesosigmoid/mesocolon/mesentery division                       & 42.03\,$\pm$\,19.37 & 21.60\,$\pm$\,31.06 & 28.54\,$\pm$\,21.15\\
S12. Sigmoid mobilisation and lateral dissection -- left            & 47.24\,$\pm$\,14.08 & 76.56\,$\pm$\,13.38 & 58.43\,$\pm$\,11.15\\
S13. Caecal mobilisation and lateral dissection -- right            & 70.83\,$\pm$\,28.93 & 14.13\,$\pm$\,29.16 & 23.57\,$\pm$\,27.08\\
S14. Lesser sac entry and/or omentum division                      & 40.37\,$\pm$\,20.48 & 50.72\,$\pm$\,31.63 & 44.96\,$\pm$\,25.56\\
S15. Splenic flexure mobilisation                                   & 27.19\,$\pm$\,37.29 & 44.89\,$\pm$\,26.64 & 33.86\,$\pm$\,24.15\\
S16. Hepatic flexure mobilisation                                   & 52.73\,$\pm$\,38.67 & 32.95\,$\pm$\,34.29 & 40.55\,$\pm$\,32.57\\
S17. Mesorectum dissection                                          & 63.75\,$\pm$\,27.46 & 63.84\,$\pm$\,18.46 & 63.80\,$\pm$\,23.18\\
S18. Distal resection site selection and preparation -- left        & 45.21\,$\pm$\,27.47 & 32.11\,$\pm$\,31.37 & 37.55\,$\pm$\,24.40\\
S19. Rectum/sigmoid transection -- left                             & 84.60\,$\pm$\,25.61 & 72.23\,$\pm$\,17.53 & 77.93\,$\pm$\,15.36\\
S20. Proximal resection site preparation and transection (intraabdominal part)            & 24.14\,$\pm$\,47.14 &  3.60\,$\pm$\,5.03  &  6.26\,$\pm$\,9.09\\
S21. Transected bowel handling and externalization (intraabdominal part)                  & 80.04\,$\pm$\,1.76  & 79.87\,$\pm$\,8.44  & 79.96\,$\pm$\,4.77\\
S22. Distal resection site preparation and transection -- right     & 90.46\,$\pm$\,4.83  & 74.20\,$\pm$\,31.66 & 81.53\,$\pm$\,35.70\\
S23. Ileal preparation and transection                              &100.00\,$\pm$\,32.69 &  3.88\,$\pm$\,36.54 &  7.46\,$\pm$\,33.64\\
S24. Dye injection and visualization                                & 98.52\,$\pm$\,3.45  & 58.38\,$\pm$\,11.08 & 73.31\,$\pm$\,7.08\\
S26. Preparation for anastomosis per rectum -- left                 & 23.14\,$\pm$\,35.73 & 45.23\,$\pm$\,17.35 & 30.62\,$\pm$\,12.97\\
S27. Rectal stump perforation and stapler firing                    & 81.67\,$\pm$\,31.73 & 32.97\,$\pm$\,32.74 & 46.98\,$\pm$\,25.25\\
S28. Leak testing -- left                                           & 46.32\,$\pm$\,12.39 & 52.65\,$\pm$\,22.13 & 49.28\,$\pm$\,3.96\\
S29. Preparation for intracorporeal anastomosis -- right            &  0.00\,$\pm$\,0.00  &  0.00\,$\pm$\,0.00  &  0.00\,$\pm$\,0.00\\
S30. Enterotomy, colotomy and intracorporeal anastomosis -- right   & 81.98\,$\pm$\,43.42 & 34.54\,$\pm$\,10.86 & 48.60\,$\pm$\,17.37\\
S31. Enterotomy/colotomy closure -- right                           & 98.49\,$\pm$\,0.00  & 74.15\,$\pm$\,2.48  & 84.61\,$\pm$\,1.53\\
S32. Additional suturing, stapling or suture removal                &  0.00\,$\pm$\,0.00 &  0.00\,$\pm$\,0.00 &  0.00\,$\pm$\,0.00\\
S33. Stoma creation (intrabdominal part)                                                 & 42.76\,$\pm$\,8.60  & 20.35\,$\pm$\,1.89  & 27.58\,$\pm$\,0.48\\
S34. Washing and aspiration, coagulation drain insertion, ports and trocar removal (intraabdominal part)         & 31.58\,$\pm$\,41.90 & 24.19\,$\pm$\,21.72 & 27.40\,$\pm$\,28.00\\\cmidrule{1-4}
\rowcolor{gray!5}\textbf{Step macro average} & \textbf{48.36\,$\pm$\,8.62} & \textbf{38.65\,$\pm$\,8.55} & \textbf{39.82\,$\pm$\,7.06}\\\bottomrule
\end{tabular}
\end{table*}

The dataset comprised 54 operative videos of five minimally invasive colorectal procedure types from four centres and one public dataset \cite{maierhein2021heico} (Table ~\ref{tab:dataset_splits}). Procedure types included left hemicolectomy (LH, n = 9), right hemicolectomy (RH, n = 13), rectal resection (RR, n = 10), sigmoid resection (SR, n = 12), and total proctocolectomy (TPC, n = 10). Mean video duration was 141\,$\pm$\,59 minutes, ranging from 93\,$\pm$\,32 minutes for right hemicolectomy to 210\,$\pm$\,50 minutes for total proctocolectomy. 

Figures \ref{fig:S1}, \ref{fig:S2} and \ref{fig:S3}  further describe the dataset: figure \ref{fig:S1} shows the frame-level composition by centre and by procedure type within each of the training, validation, and test splits; Figure \ref{fig:S2} shows per-video duration variability by procedure type, centre, and split; and Figure \ref{fig:S3} reports the normalised frame-level class distributions for phases and steps across splits, illustrating the long-tail class imbalance that motivates the use of macro-averaged F1 as the primary outcome metric. Figure \ref{fig:S4} depicts typical frames by centre and phase. 

\subsection{AI-ColoWorkflow}

The AI-ColoWorkflow model, jointly trained for phase and step recognition (see Figure \ref{fig:S5} for a comparison versus single-task training strategies), achieved a macro F1 of 73.01\%\,$\pm$\,10.27 and balanced accuracy of 73.43\% for phase recognition, and macro F1 of 39.82\%\,$\pm$\,7.06 with balanced accuracy of 38.65\% for step recognition, significantly outperforming a majority-class baseline for both phase (vs 6.71\%, $p=0.001$) and step recognition (2.29\%, $p=0.001$). Complete results are presented in Table \ref{tab:unified_results}. Figure \ref{fig:S6} shows representative AI-ColoWorkflow predictions for a video, displaying the ground-truth and predicted phase and step label timelines alongside model confidence and performance. 

At the phase level, P9 (Extracorporeal procedures) reached the highest F1 (94.73\%), followed by P2 (Vascular dissection, 76.58\%) and P5 (Anastomosis, 75.27\%). P8 (Unplanned procedures) had the lowest F1 (41.48\%). P7 (Preplanned additional procedures) was absent from the test set and excluded from evaluation. 

At the step level, the highest F1 scores were for S31 (Enterotomy/colotomy closure–right, 84.61\%), S22 (Distal transection–right, 81.53\%), and S21 (Bowel externalization, 79.96\%). S29 (Preparation for intracorporeal anastomosis – right) and S32 (Additional suturing / stapling or suture removal) scored zero F1 due to extreme rarity in the test set. S23 achieved perfect precision (100\%) but near-zero recall (3.88\%). Confusion matrices and F1 score summaries for phase and step recognition can be found in Figure \ref{fig:S7} and \ref{fig:S8}, respectively.  

Finally, Figure \ref{fig:S9} characterises AI-ColoWorkflow performance across all 10 evaluation cohorts, showing that data diversity and pooled training matter more than raw data volume. 

\subsection{AI-ColoWorkflow versus procedure- and centre-specific models }
\label{sec:comparison}

\begin{figure*}[t]
  \centering
  \includegraphics[width=0.80\textwidth, trim={0mm 30mm 10mm 10mm}, clip]{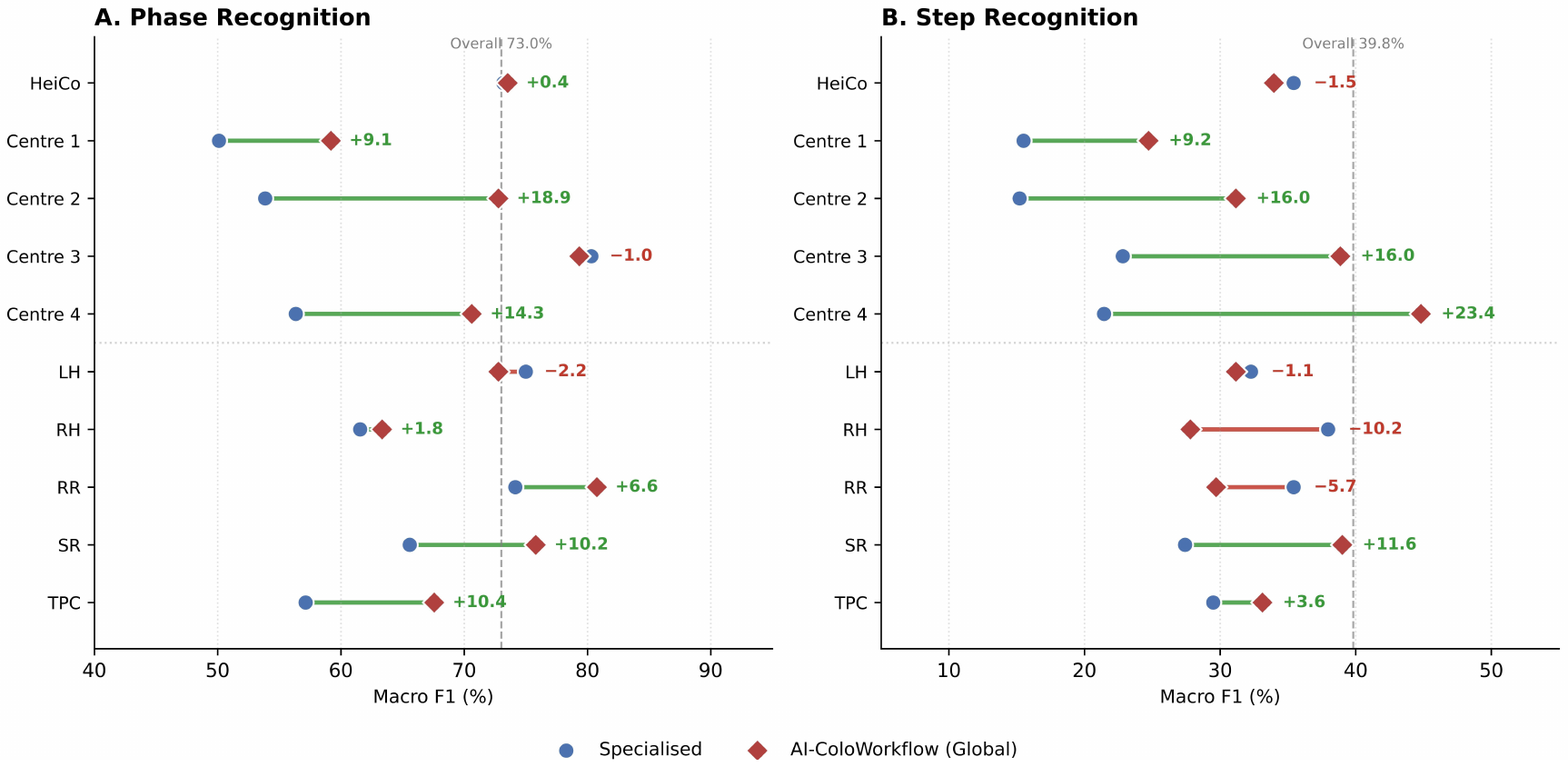}
  \caption{Comparison of AI-ColoWorkflow performance versus procedure and centre-specific models. Numbers and *** in green suggest superiority of AI-ColoWorkflow while those in red suggest superiority of specific models. LH: Left hemicolectomy; RH: Right hemicolectomy; RR: Rectal resection; SR: Sigmoid resection; TPC: Total proctocolectomy.}
  \label{fig:fig2}
\end{figure*}

Figure \ref{fig:fig2} plots the performance of AI-ColoWorkflow versus procedure and centre specific models’ performance; see Table \ref{tab:comparison} for complete F1 score comparison. For phase recognition, AI-ColoWorkflow outperformed centre-specific models in 4/5 centres (largest margins: Centre 2 +18.9, Centre 4 +14.3\%; Centre 3 was the exception at -1.0) and procedure-specific models in 4/5 types (TPC +10.4, SR +10.2, RR +6.6, RH +1.8; LH was the exception at -2.2). For step recognition, the AI-ColoWorkflow model outperformed centre-specific models in 4/5 centres (Centre 4 +23.4, Centre 3 +16.0, Centre 2 +15.9, Centre 1 +9.2; HeiCo was the exception at -1.5) and procedure-specific models in 2/5 types (SR +11.6, TPC +3.6). RH showed the largest step advantage for the procedure-specific model (-10.2\%), consistent with its more constrained step ontology. 

\subsection{Generalization}

Leave-one-centre-out (LOCO) cross-validation F1 scores are reported in Table ~\ref{tab:loco}, Table \ref{tab:loco_full} reports all computed metrics, Figure \ref{fig:S10} shows performance heatmaps. 

Mean phase F1 on the held-out centre was 48.42\% (range: 37.49\% for Centre 2 to 65.78\% for Centre 4). Performance on the four seen centres was stable across folds (mean 61.59\% phase, 40.61\% step). Centre 4 showed the smallest LOCO penalty for phase (held-out 65.78\%, AI-ColoWorkflow 73.01\%; gap -7.2\%), likely owing to the homogeneous single-procedure composition of that centre. Centre 2 showed the largest held-out phase performance drop (37.49\%, retention 51.3\%), while Centre 1 and HeiCo also showed substantial drops, consistent with their smaller and more heterogeneous contributions to the pooled training set. 

\begin{table}[h]
\centering
\caption{Leave-one-centre-out (LOCO) cross-validation results. Macro F1 (\%) on the held-out centre and the remaining seen centres per fold, compared with the globally trained AI-ColoWorkflow model. All values are macro averaged.}
\label{tab:loco}
\small
\setlength{\tabcolsep}{4.5pt}
\begin{tabular}{@{}lcccc@{}}
\toprule
& \multicolumn{2}{c}{\textbf{Phase F1 (\%)}} & \multicolumn{2}{c}{\textbf{Step F1 (\%)}}\\\cmidrule(lr){2-3}\cmidrule(lr){4-5}
\textbf{Held-out} & \textbf{Held-out} & \textbf{Seen} & \textbf{Held-out} & \textbf{Seen}\\\midrule
HeiCo      & 43.83 & 65.10 & 16.74 & 40.44\\
Centre 1      & 43.46 & 60.63 & 18.57 & 39.28\\
Centre 2       & 37.49 & 60.05 & 23.38 & 42.24\\
Centre 3    & 51.56 & 60.62 & 18.11 & 36.67\\
Centre 4 & 65.78 & 61.55 & 32.91 & 44.42\\\midrule
\rowcolor{gray!8}\textbf{Mean} & \textbf{48.42} & \textbf{61.59} & \textbf{21.94} & \textbf{40.61}\\\midrule
\multicolumn{5}{l}{\textit{Global model (all centres in training):}}\\
\rowcolor{green!8}\textbf{Overall} & \multicolumn{2}{c}{\textbf{73.01}} & \multicolumn{2}{c}{\textbf{39.82}}\\\bottomrule
\end{tabular}
\end{table}

\section{Discussion}

This study developed and evaluated AI-ColoWorkflow, to the best of our knowledge, the first AI model designed to perform automated workflow analysis across multiple laparoscopic and robotic colorectal procedure types using a unified, consensus-based ontology.  

AI-ColoWorkflow achieved a macro F1 score of 73.01\% for phase recognition and 39.82\% for step recognition across five procedure types and five centres. Systematic comparison with procedure- and centre-specific models demonstrated that a single model jointly trained on diverse, pooled data consistently outperforms localised models for phase recognition in four of five procedure types and four of five centres, and outperforms specialised models for step recognition in four of five centres but only two of five procedure types. These results suggest that pooled training is a broadly effective strategy for phase-level recognition and centre generalisation, while procedure-specific step models retain selective advantages for procedures with constrained, visually distinct steps (RH, -10.2\%for step; RR, -5.7\%). The leave-one-centre-out analysis revealed a generalisation gap across all centres, with phase macro F1 on the held-out centre ranging from 37.49\% (Centre 2) to 65.78\% (Centre 4), corresponding to performance retention of 51.3\% to 90.1\% relative to the globally trained model. These findings collectively establish an empirical baseline for cross-procedural AI-based workflow analysis in colorectal surgery and provide quantitative evidence on the trade-offs between global and specialised training strategies.

AI-ColoWorkflow is grounded in the expert-consensus ColoWorkflow hierarchical annotation framework \cite{jain2026video}, comprising procedure-agnostic phases and procedure-specific steps, consistent with the SAGES consensus recommendations on surgical video annotation \cite{meireles2021sages}. This hierarchy enables meaningful comparison across procedure types at the phase level while preserving granularity at the step level. Prior AI models for workflow recognition in colorectal surgery have been trained and evaluated on single procedure types, including sigmoid resection \cite{nakajima2024automated} and rectal surgery \cite{kolbinger2023context}, or on procedure-specific phase definitions derived from individual datasets such as HeiCo \cite{wagner2023heichole}. In laparoscopic cholecystectomy, where workflow recognition is most mature, models trained on the Cholec80 dataset achieve phase-level accuracy above 90\% \cite{liu2025lovit}, but this procedure is substantially shorter, more standardised, and defined by fewer phases. The lower absolute performance observed in the present study reflects the greater complexity of the task: more procedure types, more phases and steps, higher inter- and intra-procedural variability, and longer operative times. Notably, Marginal Gains studies  \cite{moralesconde2025marginal,moralesconde2026shortterm} applied AI-enhanced video-based assessment to laparoscopic right hemicolectomy using a procedure-specific workflow, demonstrating the feasibility of integrating AI and VBA in colorectal surgical training. AI-ColoWorkflow extends this approach by applying a single, unified model across the full spectrum of minimally invasive colorectal procedure types. 

The observation that the global model generally outperformed specialised models is consistent with findings from multi-institutional studies of surgical workflow recognition, where data diversity during training improves cross-site robustness \cite{lavanchy2024challenges,alekseenko2026genguard}. This advantage was most pronounced at centres and procedure types with smaller local datasets–Centre 1 (+9.1\% phase, +9.2\% step), Centre 2 (+18.9\% phase, +15.9\% step), and SR (+10.2\% phase, +11.6\% step)  -- consistent with the hypothesis that pooled training compensates for limited institutional volume by providing a richer and more varied set of training examples. The principal exceptions were Centre 3 (phase -1.0\% ) and RH and RR (step -10.2\% and -5.7\% respectively), where specialised models performed better. For Centre 3, the global model matched the specialist closely (a -1.0\% margin within noise). For RH and RR, the step-level advantage of the specialised models most likely reflects these procedures’ more constrained and visually predictable step sequences, where the additional diversity introduced by pooled training may act as noise rather than a regulariser. The LOCO analysis provided complementary evidence: Centre 4 retained 90.1\% of global phase performance when trained without its own data, suggesting that its sigmoid resection videos share substantial visual and procedural overlap with sigmoid resection videos from other centres. Conversely, Centre 2 retained only 51.3\%, and Centre 1 only 59.5\%, underscoring that the global model’s performance at these sites is heavily dependent on the presence of local training examples–a finding with direct implications for the adoption of AI-ColoWorkflow at new, data-poor institutions. 

The performance gap between phase and step recognition reflects fundamental differences in task structure as well as data distribution, and it is consistent with previous findings \cite{lavanchy2024challenges}. Phase recognition benefited from longer, visually distinct temporal segments and a smaller evaluated label set (8 classes), providing the model with more training signal per class and reducing the risk of ambiguous transitions. Step recognition, by contrast, required discrimination between brief, visually similar actions occurring within the same anatomical field (33 evaluated classes), many of which differ only in subtle instrument movements or tissue handling cues that are difficult to distinguish at 1 frame per second. Steps with high performance, such as enterotomy/colotomy closure (S31, F1 84.61\%), right-sided distal transection (S22, F1 81.53\%), and bowel externalisation (S21, F1 79.96\%) correspond to actions with visually distinctive signatures, including specific suturing or stapling motions and the characteristic exteriorisation manoeuvre. Steps with low or zero performance, such as proximal resection site preparation (S20, F1 6.26\%), ileal preparation (S23, F1 7.46\%), and intracorporeal anastomosis preparation (S29, F1 0\%), lack consistent anatomical landmarks, are visually similar to adjacent steps, and are severely underrepresented in the 54-video dataset. This long-tail distribution of surgical actions, in which clinically important but infrequent events are too rare to be reliably learned from small datasets, represents a fundamental challenge for step-level recognition that is not solvable by architectural choices alone and will require substantially larger annotated datasets.

This study has several limitations. The 54-video dataset, while multicentre and multi-procedural, remains limited in absolute size, particularly for rare steps that appear in only a subset of procedure types. This is a common constraint in Surgical Data Science, where video collection, de-identification, and expert annotation are resource intensive. The class imbalance is especially pronounced for short-duration steps, which in some cases had fewer training frames than test frames, making reliable model learning impossible and complicating fair evaluation. Annotations were performed by a single medical doctor rather than multiple independent raters; however, this annotator was extensively trained in VBA methodology, having participated directly in the development and validation of the ColoWorkflow framework and its annotation guidelines \cite{jain2026video}, which mitigates but does not eliminate potential annotation bias. The study was limited to the temporal dimension of surgical workflow (phase and step recognition) and did not address spatial tasks such as instrument detection, anatomy recognition, or critical view of safety assessment, which represent complementary and clinically relevant avenues for future work. 

Future studies should prioritise expanding the dataset through multi-institutional collaboration, with particular attention to underrepresented procedure types and rare surgical steps where current performance is limited by data volume rather than model capacity. The integration of additional data modalities, such as instrument tracking signals from robotic platforms, device data from the operating room, or audio signals–could improve recognition of visually ambiguous steps that are currently indistinguishable from video alone. Prospective validation in clinical settings is needed to assess whether AI-ColoWorkflow can reduce the time burden of manual annotation at scale and support real-time intraoperative workflow monitoring. Robustness to domain shift, including different camera systems, lighting conditions, and robotic versus laparoscopic approaches, should be explicitly characterised in future generalisability studies. Finally, the relationship between AI-detected workflow patterns and clinical outcomes, such as operative time, complication rates, and learning curves, remains to be investigated and represents a necessary step toward demonstrating the clinical value of automated surgical workflow analysis beyond its utility as a research tool. 

\noindent\textbf{Sources of Funding:} This work has received funding from the European Union (ERC, CompSURG, 101088553). Views and opinions expressed are, however, those of the authors only and do not necessarily reflect those of the European Union or the European Research Council. Neither the European Union nor the granting authority can be held responsible for them. This work was partially supported by French state funds managed by the ANR under Grant ANR-10-IAHU-02 (IHU Strasbourg) and ANR-22-FAI1-0001 (project DAIOR). 

\noindent\textbf{Conflict of Interest Statement:} Pietro Mascagni and Nicolas Padoy are co-founders and shareholders of Scialytics. All other authors declare no conflicts of interest.

\bibliographystyle{elsarticle-num}
\bibliography{sn-bibliography}

\clearpage
\appendix
\onecolumn
\section*{Supplementary Figures and Tables}

\setcounter{figure}{0}
\setcounter{table}{0}
\renewcommand{\thefigure}{S\arabic{figure}}
\renewcommand{\thetable}{S\arabic{table}}

\begin{table}[h]
\centering
\caption{Hardware and training configuration.}
\label{tab:supp_specs}
\small
\setlength{\tabcolsep}{8pt}
\begin{tabular}{@{}ll@{}}
\toprule
\textbf{Component} & \textbf{Specification}\\\midrule
\multicolumn{2}{l}{\textit{Hardware}}\vspace{2pt}\\
GPU & NVIDIA V100 (32\,GB)\\\midrule
\multicolumn{2}{l}{\textit{Spatial encoder}}\vspace{2pt}\\
Architecture & DINOv3 vision transformer (fine-tuned)\\
Input resolution & 224x224\\\midrule
\multicolumn{2}{l}{\textit{Temporal model}}\vspace{2pt}\\
Architecture & SAHC (Segment-Attentive Hierarchical Consistency TCN)\\
Training mode & Joint (shared trunk, two output heads: phase \& step)\\
Optimiser & Adam ($\beta_1=0.9$, $\beta_2=0.999$)\\
Weight decay & $1\times10^{-4}$\\\midrule
\multicolumn{2}{l}{\textit{Libraries}}\vspace{2pt}\\
Framework & PyTorch \cite{paszke2019pytorch}\\
Frame processing & OpenCV \cite{bradski2000opencv}\\
Evaluation metrics & scikit-learn \cite{pedregosa2011scikit}\\\bottomrule
\end{tabular}
\end{table}

\begin{figure}[H]
  \centering
  \includegraphics[width=0.96\textwidth, trim={0mm 25mm 5mm 25mm}, clip]{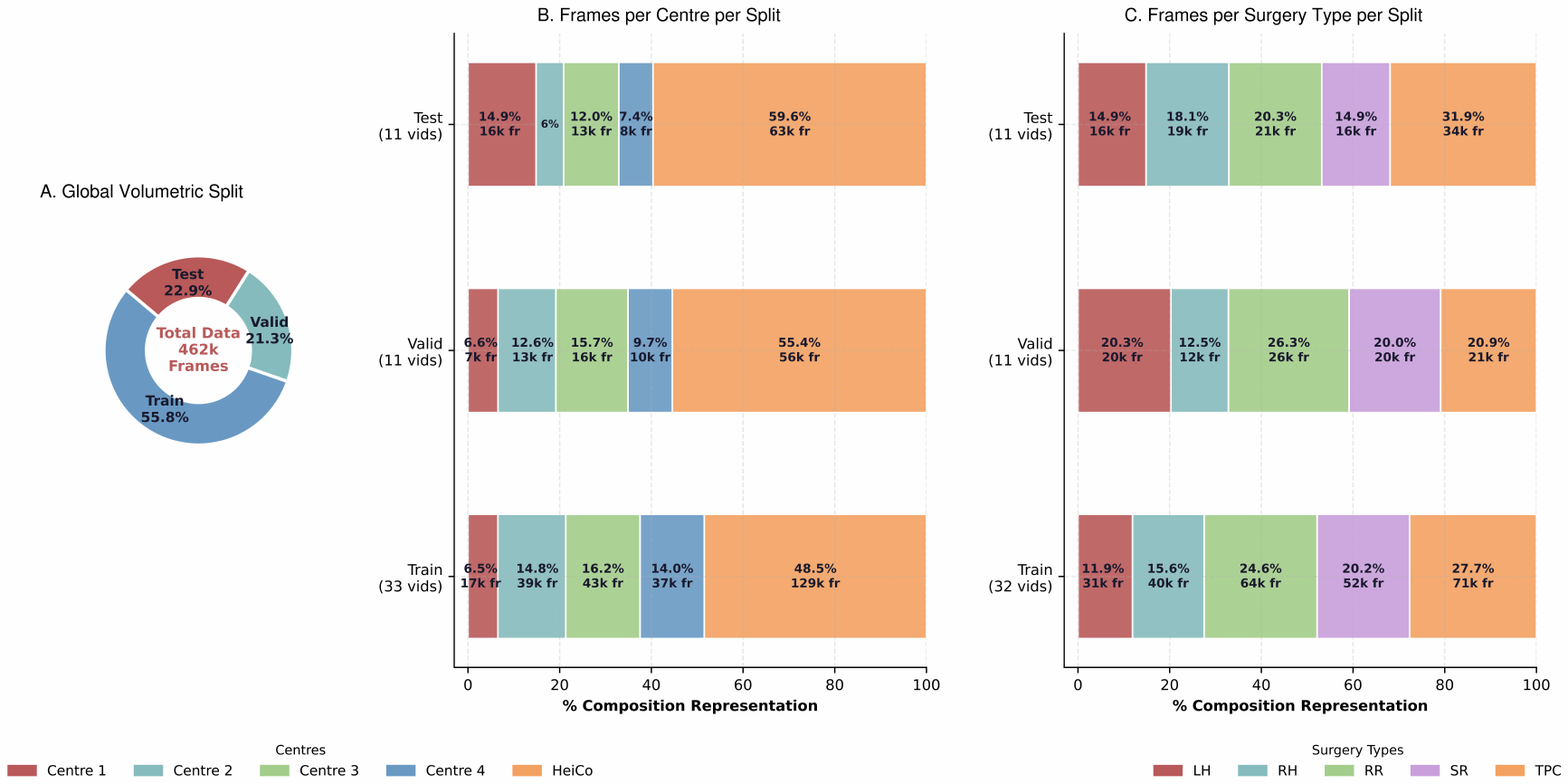}
  \caption{Frame-level dataset composition by centre and surgery type within each of the training, validation, and test splits. (A) Donut chart showing overall train/validation/test frame proportions (462k total frames in total). (B) Stacked horizontal bars showing the percentage of frames contributed by each centre within each split. (C) Stacked bars showing the percentage of frames contributed by each surgery type within each split. Fr: frames; LH: Left hemicolectomy; RH: Right hemicolectomy; RR: Rectal resection; SR: Sigmoid resection; TPC: Total proctocolectomy.}
  \label{fig:S1}
\end{figure}

\begin{figure}[H]
  \centering
  \includegraphics[width=0.96\textwidth, trim={0mm 55mm 5mm 35mm}, clip]{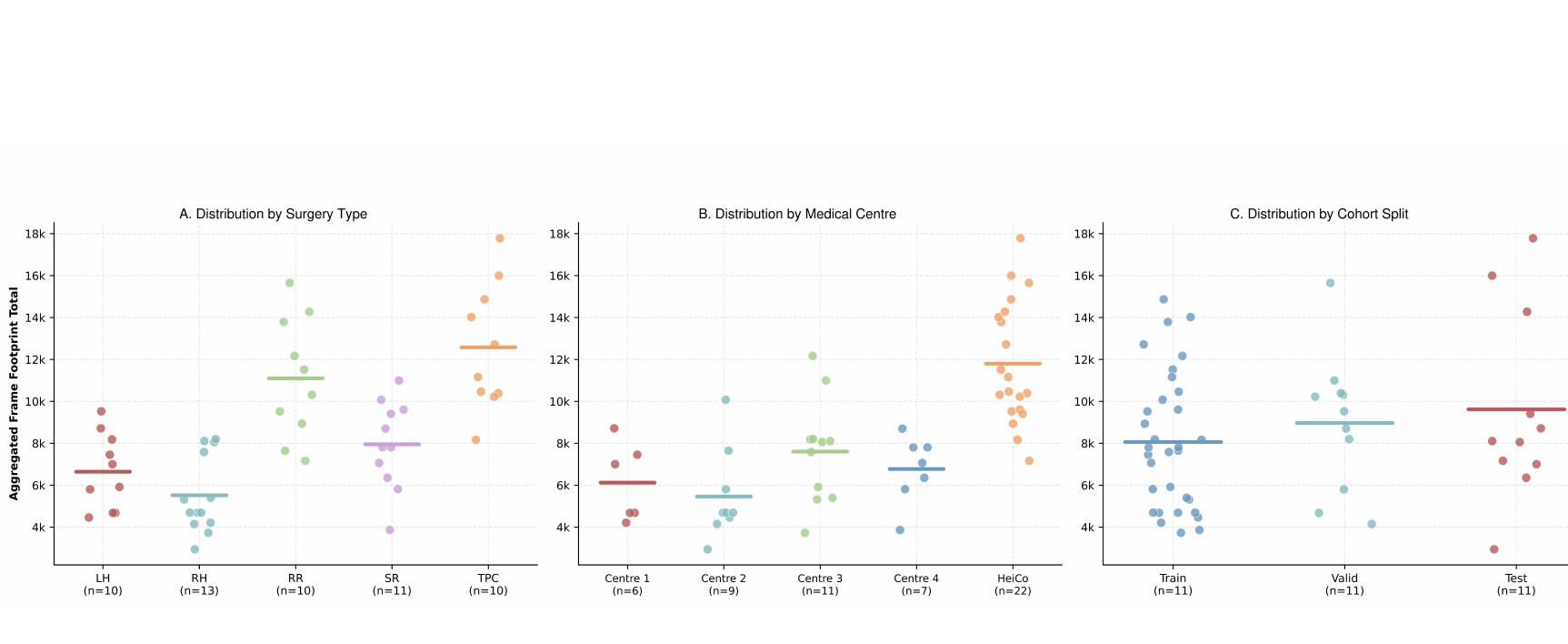}
  \caption{Per-video duration variability by procedure type, centre, and split at 1 frame per second. Strip plots showing individual video frame counts grouped by (A) surgery type, (B) centre, and (C) split. Horizontal bars mark group means. LH: Left hemicolectomy; RH: Right hemicolectomy; RR: Rectal resection; SR: Sigmoid resection; TPC: Total proctocolectomy.}
  \label{fig:S2}
\end{figure}

\begin{figure}[H]
  \centering
  \includegraphics[width=0.90\textwidth, trim={0mm 10mm 5mm 25mm}, clip]{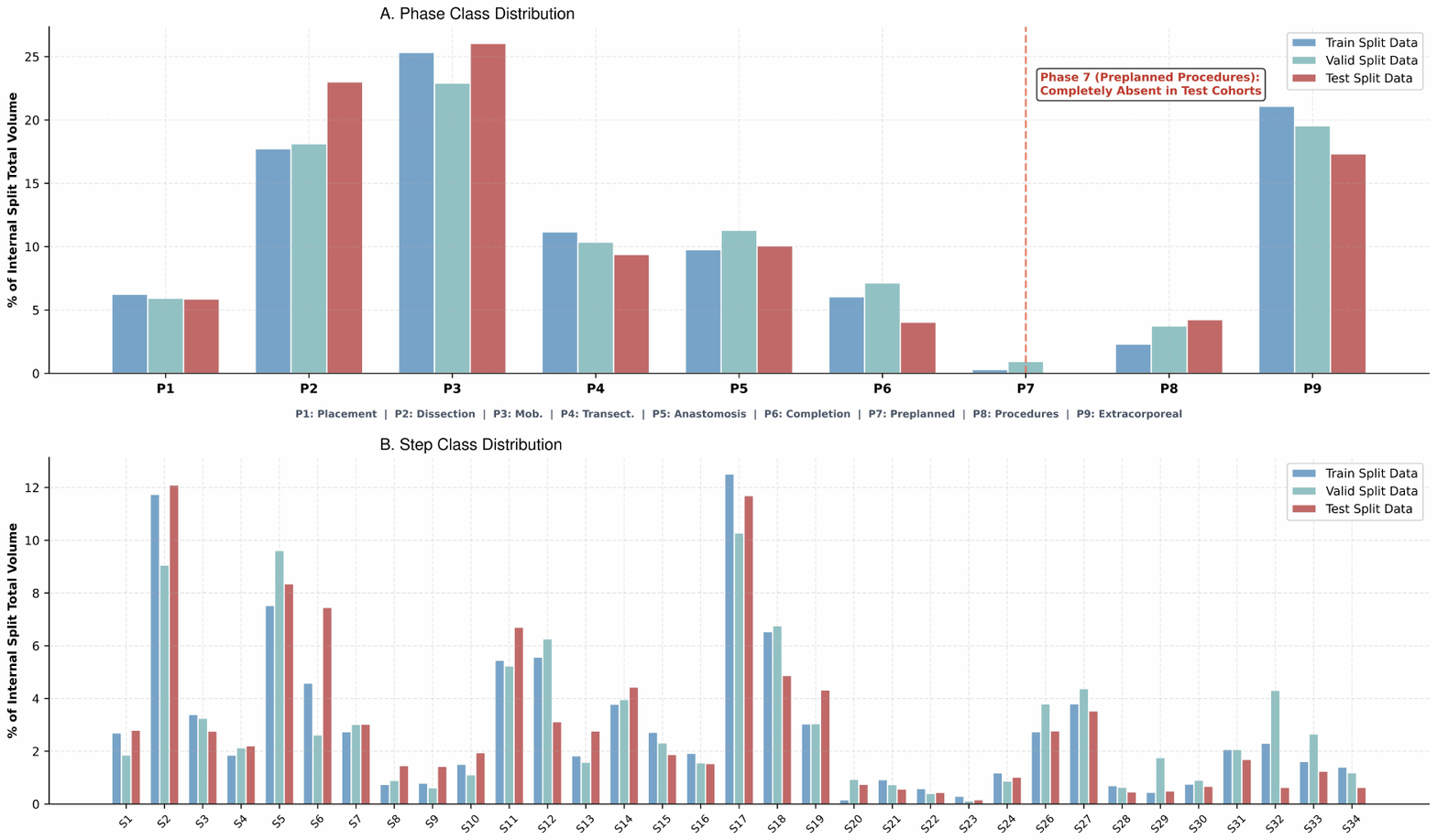}
  \caption{Normalised frame-level class distributions for phases and steps across splits. (A) Grouped bar chart of phase frame counts normalised by split total (\% of split frames) for P1–P9; P7 (Preplanned additional procedures) is absent from the test set and was excluded from evaluation. (B) Grouped bar chart of normalised step frame counts (\% of split frames) for all 33 evaluated steps (S25 is an unused gap), illustrating the long-tail imbalance that motivates macro-averaged F1 as the primary metric. Complete phases (P) and steps (S) definitions can be found in Table \ref{tab:unified_results}.}
  \label{fig:S3}
\end{figure}

\begin{figure}[H]
  \centering
  \includegraphics[width=\textwidth, trim={0mm 50mm 5mm 20mm}, clip]{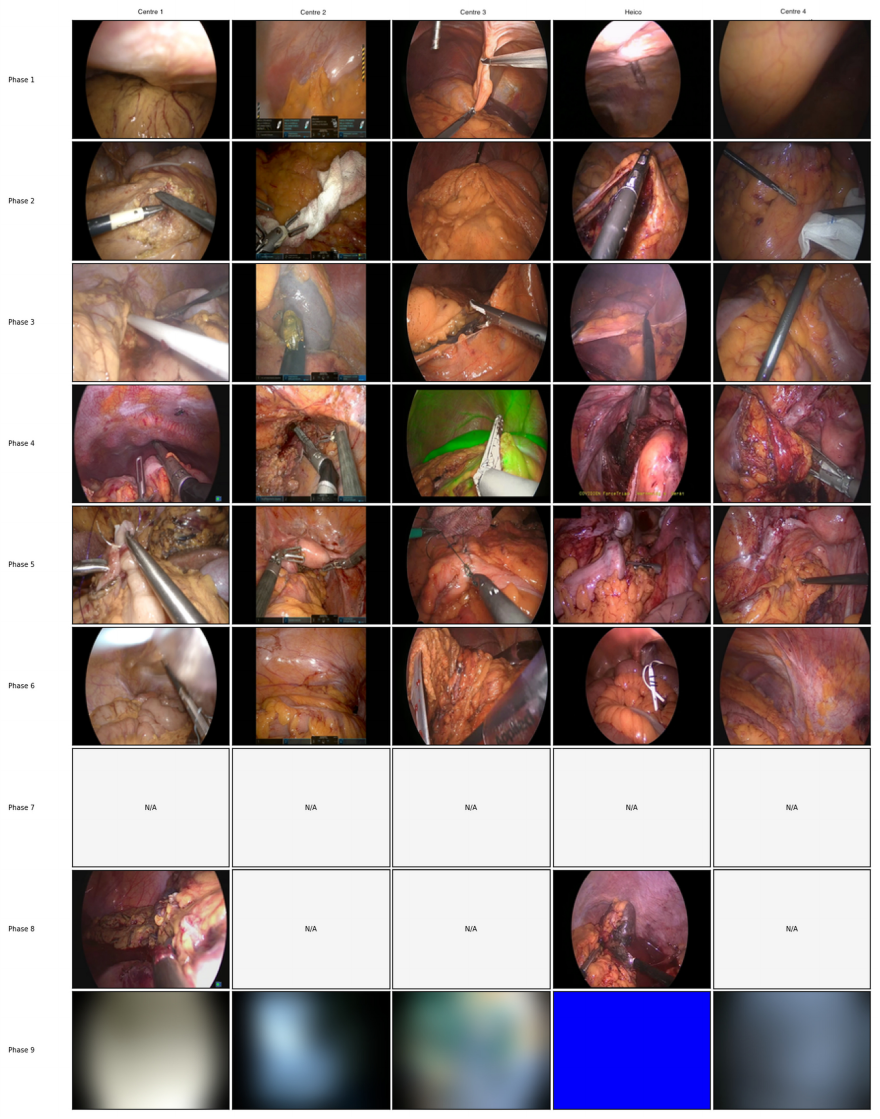}
  \caption{Typical frames by centre and phase. Each cell shows a representative frame for the (centre, phase) pair. Phase 9 (Extracorporeal procedures) is blurred to comply with institutional privacy requirements; grey cells indicate no available frames.}
  \label{fig:S4}
\end{figure}

\begin{figure}[H]
  \centering
  \includegraphics[width=0.80\textwidth, trim={10mm 5mm 10mm 5mm}, clip]{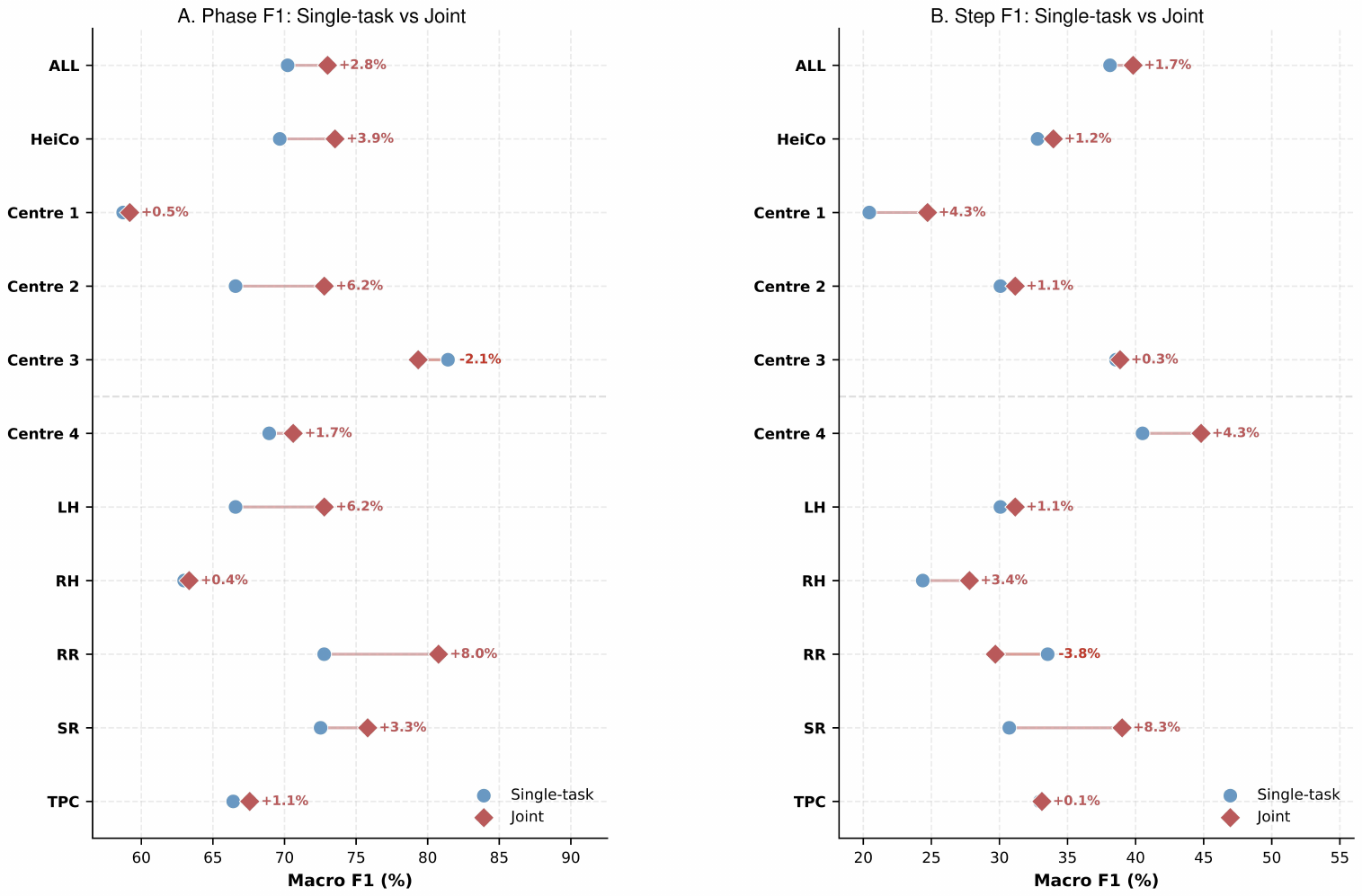}
  \caption{Single-task versus joint training. Joint training improved macro F1 by +2.8\% for phase (70.2\% to 73.0\%) and +1.7\% for step (38.1\% to 39.8\%) relative to separately trained single-task models, with gains consistent across nearly all subgroups. (A) Phase recognition; (B) step recognition. Each row is a cohort (overall, 5 centres, 5 procedure types). Circles = single-task model F1; diamonds = joint model (AI-ColoWorkflow) F1. Green connectors indicate joint training improved performance; red indicate regression. Annotations show $\Delta$F1 (joint $-$ single-task).}
  \label{fig:S5}
\end{figure}

\begin{figure}[H]
  \centering
  \includegraphics[width=0.80\textwidth, trim={5mm 20mm 5mm 25mm}, clip]{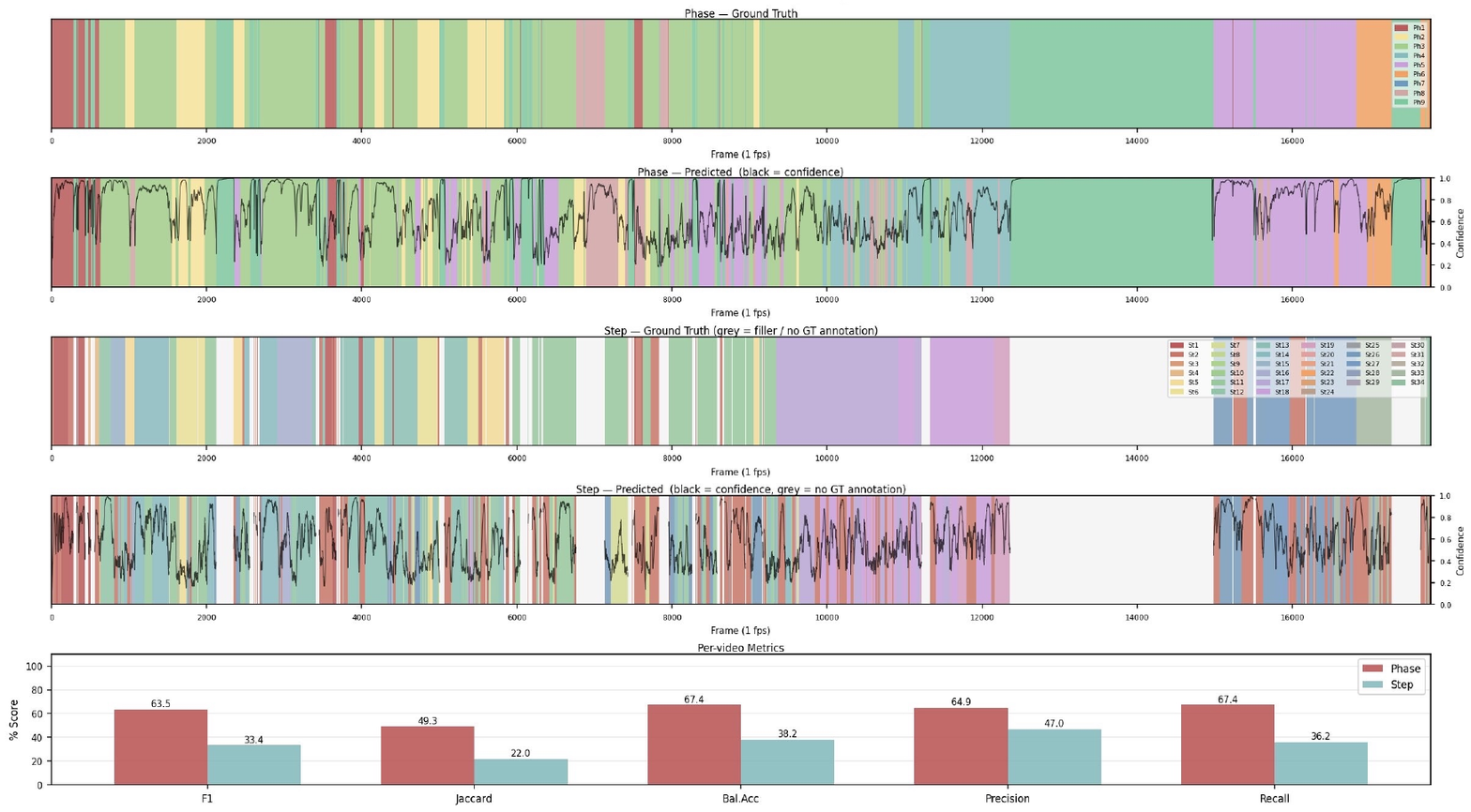}
  \caption{Predictions for a representative video. (A) Ground-truth phase timeline; (B) predicted phase timeline, with model confidence (softmax maximum probability, black line); (C) ground-truth step timeline; (D) predicted step timeline with model confidence, grey regions indicate frames with no step ground-truth annotation; (E) per-video F1, Jaccard, balanced accuracy, precision, and recall.}
  \label{fig:S6}
\end{figure}

\begin{figure}[H]
  \centering
  \includegraphics[width=0.90\textwidth, trim={0mm 30mm 5mm 15mm}, clip]{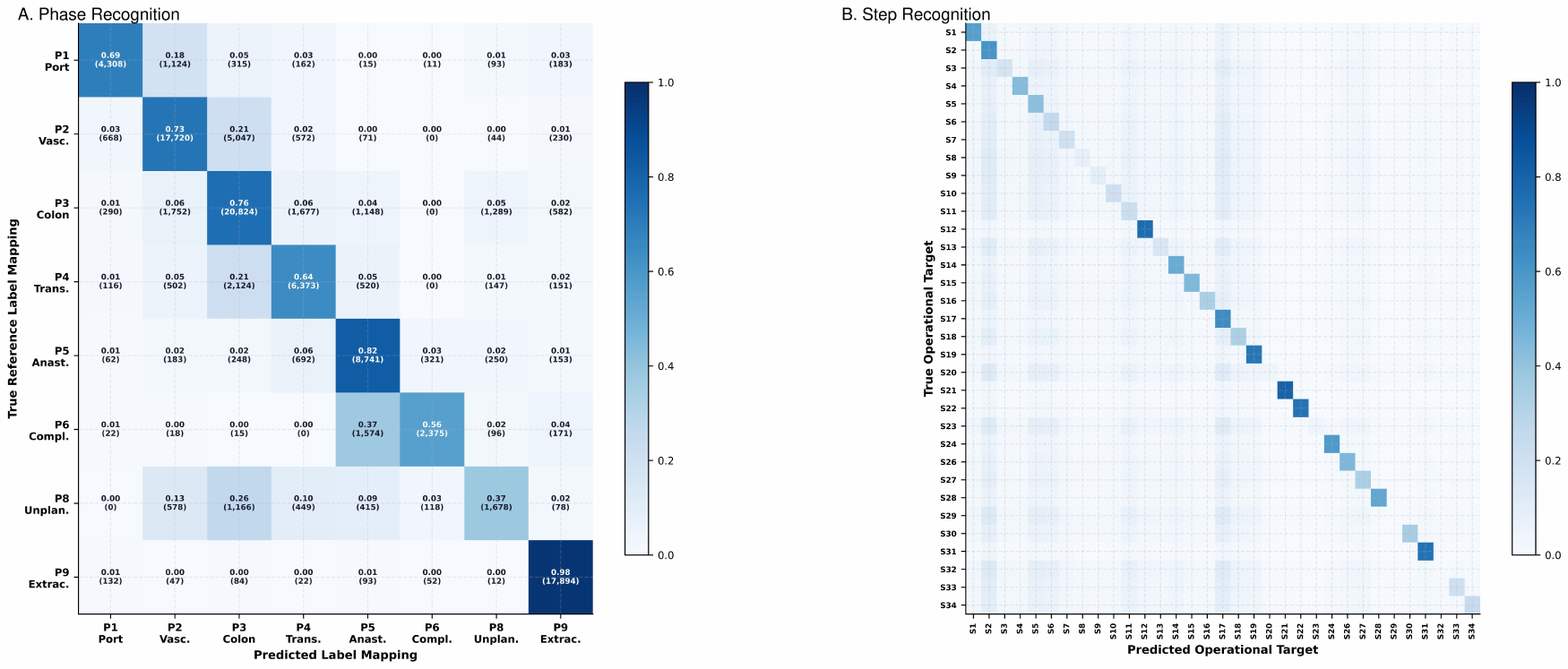}
  \caption{Confusion matrices for phase (A) and step (B) recognition. Row-normalised matrices: each cell shows the fraction of true-class frames predicted as each target class; diagonal = correct prediction. The phase confusion matrix is computed from model outputs; the step confusion matrix is approximated from per-class recall and support (i.e., the number of samples). P8 (Unplanned procedures) shows the most diffuse phase confusion. Step classes with near-zero recall (S20, S23, S29, S32) appear as near-empty rows.}
  \label{fig:S7}
\end{figure}

\begin{figure}[H]
  \centering
  \includegraphics[width=0.90\textwidth, trim={5mm 25mm 5mm 40mm}, clip]{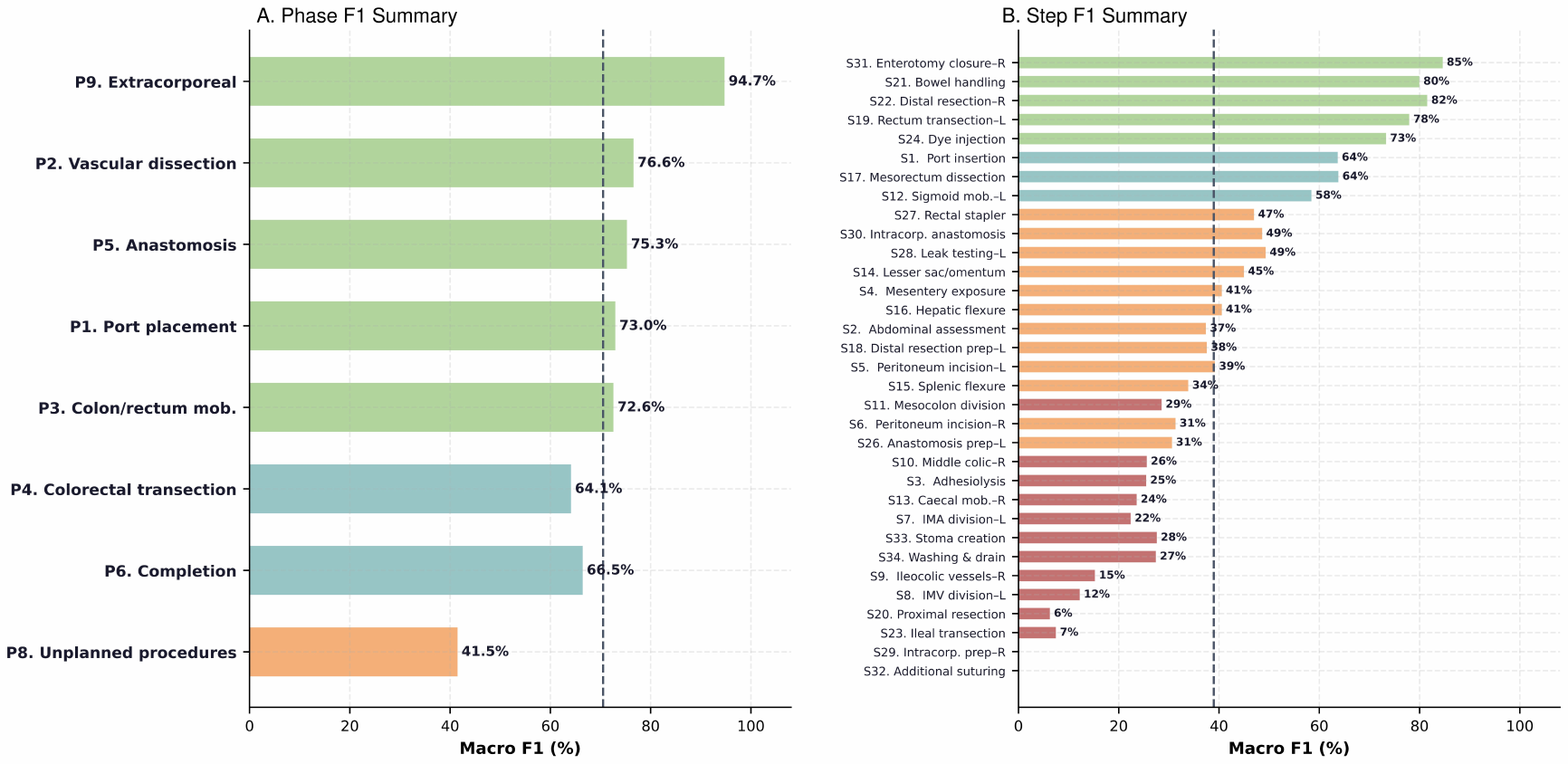}
  \caption{Phase and step F1 score summary. (A) Phase classes (n = 8 evaluated); (B) step classes (n = 33 evaluated), sorted by F1 in descending order. Bars are coloured by F1 tier (green $\geq$70\%, blue 50--70\%, orange 30--50\%, red <30\%). The dashed vertical line marks the macro mean.}
  \label{fig:S8}
\end{figure}

\begin{figure}[H]
  \centering
  \includegraphics[width=0.90\textwidth]{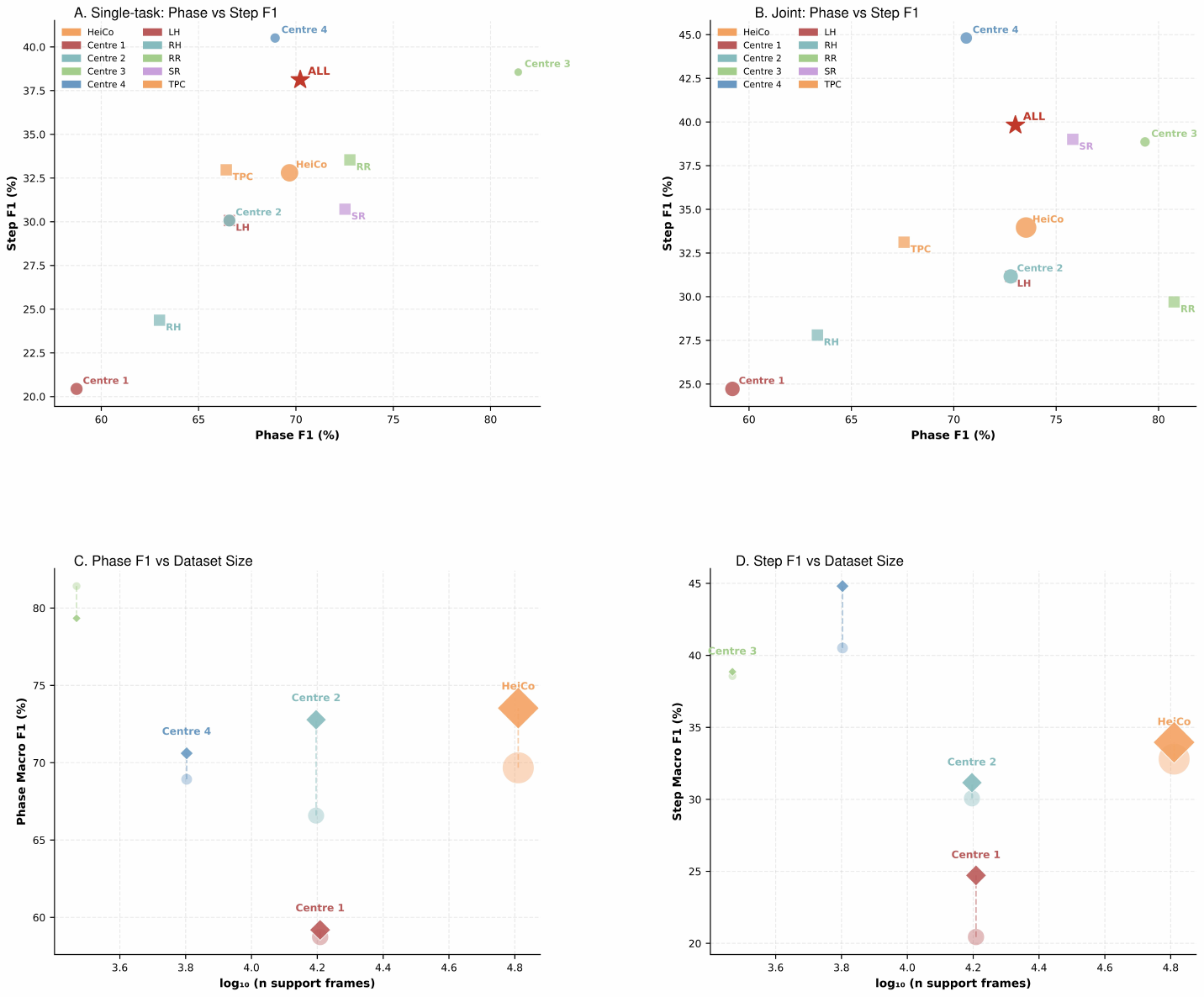}
  \caption{AI-ColoWorkflow performance across evaluation cohorts. (A) Single-task (i.e., phase or step recognition) training: scatter of phase against step macro F1 score. (B) Joint (i.e., phase and step recognition) training: scatter of phase against step macro F1 score. For A and B, circles represent centres (sized by frame count) while squares represent procedure types, star represents overall. (C) Phase macro F1 versus log-transformed frame count per centre. (D) Step macro F1 versus log-transformed frame count per centre. For C and D, circles represent single-task training while diamonds represent joint training; areas are proportional to frame count.}
  \label{fig:S9}
\end{figure}

\begin{table}[h]
\centering
\caption{AI-ColoWorkflow (global) versus procedure- and centre-specific models. Macro-averaged F1 score (\%) on the test set. $\Delta$ = global - specialised. LH: Left hemicolectomy; RH: Right hemicolectomy; RR: Rectal resection; SR: Sigmoid resection; TPC: Total proctocolectomy.}
\label{tab:comparison}
\footnotesize
\setlength{\tabcolsep}{5.5pt}
\begin{tabular}{@{}llcccccc@{}}
\toprule
& & \multicolumn{3}{c}{\textbf{Phase F1 (\%)}} & \multicolumn{3}{c}{\textbf{Step F1 (\%)}}\\\cmidrule(lr){3-5}\cmidrule(lr){6-8}
\textbf{Cohort} & \textbf{Type} & \textbf{Global} & \textbf{Spec.} & \bm{$\Delta$} & \textbf{Global} & \textbf{Spec.} & \bm{$\Delta$}\\\midrule
HeiCo      & Centre & 73.53 & 73.18 & \win{0.4}  & 33.96 & 35.42 & \loss{1.5}\\
Centre 1      & Centre & 59.19 & 50.11 & \win{9.1}  & 24.72 & 15.50 & \win{9.2}\\
Centre 2       & Centre & 72.78 & 53.86 & \win{18.9} & 31.16 & 15.21 & \win{15.9}\\
Centre 3    & Centre & 79.34 & 80.31 & \loss{1.0} & 38.86 & 22.82 & \win{16.0}\\
Centre 4 & Centre & 70.61 & 56.34 & \win{14.3} & 44.81 & 21.44 & \win{23.4}\\\midrule
LH  & Type & 72.78 & 75.00 & \loss{2.2} & 31.16 & 32.27 & \loss{1.1}\\
RH  & Type & 63.34 & 61.56 & \win{1.8}  & 27.80 & 37.96 & \loss{10.2}\\
RR  & Type & 80.76 & 74.15 & \win{6.6}  & 29.70 & 35.42 & \loss{5.7}\\
SR  & Type & 75.81 & 65.58 & \win{10.2} & 39.01 & 27.41 & \win{11.6}\\
TPC & Type & 67.57 & 57.14 & \win{10.4} & 33.12 & 29.49 & \win{3.6}\\\midrule
\rowcolor{gray!5}\textbf{Overall} & & \textbf{73.01} & --- & --- & \textbf{39.82} & --- & ---\\\bottomrule
\end{tabular}
\end{table}

\begin{table}[h]
\centering
\caption{Full leave-one-centre-out (LOCO) cross-validation results. For each fold (held-out centre), metrics are reported separately on the held-out centre and on the four seen centres. All values are macro-averaged (\%).}
\label{tab:loco_full}
\footnotesize
\setlength{\tabcolsep}{4pt}
\begin{tabular}{@{}llccccc@{}}
\toprule
\textbf{Held-out centre} & \textbf{Evaluation on} & \textbf{Task} & \textbf{Balanced accuracy (\%)} & \textbf{F1 (\%)} & \textbf{Precision (\%)} & \textbf{Recall (\%)}\\\midrule
\multirow{4}{*}{HeiCo}
  & \multirow{2}{*}{Held-out} & Phase & 48.01 & 43.83 & 45.51 & 48.01\\
  &                            & Step  & 23.42 & 16.74 & 20.88 & 18.45\\\cmidrule(lr){2-7}
  & \multirow{2}{*}{Seen centres} & Phase & 66.12 & 65.10 & 64.54 & 66.12\\
  &                            & Step  & 40.37 & 40.44 & 51.15 & 40.37\\\midrule
\multirow{4}{*}{Centre 1}
  & \multirow{2}{*}{Held-out} & Phase & 45.32 & 43.46 & 47.50 & 45.32\\
  &                            & Step  & 21.99 & 18.57 & 29.44 & 21.31\\\cmidrule(lr){2-7}
  & \multirow{2}{*}{Seen centres} & Phase & 60.09 & 60.63 & 62.16 & 60.09\\
  &                            & Step  & 37.58 & 39.28 & 55.56 & 37.58\\\midrule
\multirow{4}{*}{Centre 2}
  & \multirow{2}{*}{Held-out} & Phase & 53.94 & 37.49 & 45.56 & 41.95\\
  &                            & Step  & 30.92 & 23.38 & 32.98 & 29.99\\\cmidrule(lr){2-7}
  & \multirow{2}{*}{Seen centres} & Phase & 58.38 & 60.05 & 62.82 & 58.38\\
  &                            & Step  & 39.58 & 42.24 & 54.00 & 39.58\\\midrule
\multirow{4}{*}{Centre 3}
  & \multirow{2}{*}{Held-out} & Phase & 54.46 & 51.56 & 55.49 & 54.46\\
  &                            & Step  & 22.63 & 18.11 & 32.91 & 21.93\\\cmidrule(lr){2-7}
  & \multirow{2}{*}{Seen centres} & Phase & 60.04 & 60.62 & 61.75 & 60.04\\
  &                            & Step  & 36.78 & 36.67 & 49.29 & 36.78\\\midrule
\multirow{4}{*}{Centre 4}
  & \multirow{2}{*}{Held-out} & Phase & 64.78 & 65.78 & 68.75 & 64.78\\
  &                            & Step  & 40.22 & 32.91 & 40.21 & 34.03\\\cmidrule(lr){2-7}
  & \multirow{2}{*}{Seen centres} & Phase & 60.55 & 61.55 & 63.81 & 60.55\\
  &                            & Step  & 43.34 & 44.42 & 50.21 & 43.34\\\midrule
\rowcolor{green!8}
  &
  & Phase & \textbf{73.43} & \textbf{73.01} & \textbf{73.19} & \textbf{73.43}\\
\rowcolor{green!8}
\multirow{-2}{*}{\textbf{Global model}} & \multirow{-2}{*}{\textbf{All centres}}
  & Step & \textbf{38.65} & \textbf{39.82} & \textbf{48.36} & \textbf{38.65}\\\bottomrule
\end{tabular}
\end{table}

\begin{figure}[H]
  \centering
  \includegraphics[width=0.90\textwidth, trim={5mm 50mm 5mm 45mm}, clip]{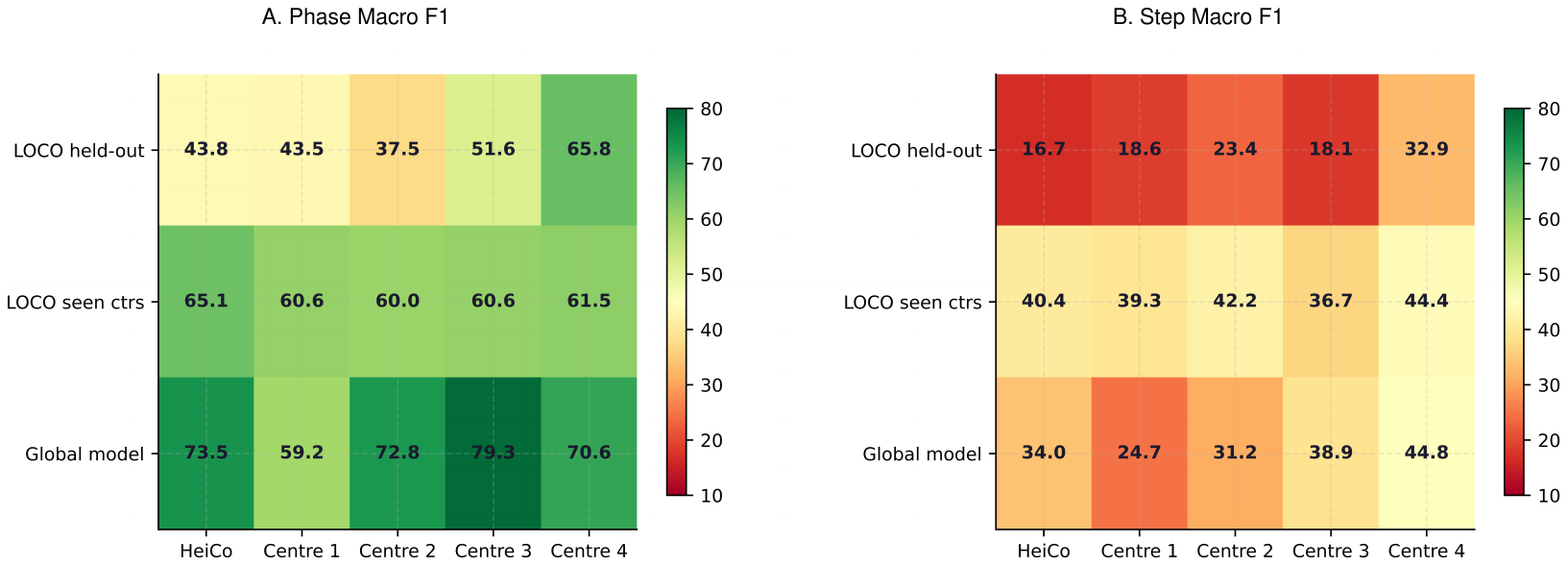}
  \caption{Leave-one-centre-out (LOCO) performance heatmaps for phase (A) and step (B) recognition. Three conditions (rows): LOCO held-out, LOCO seen centres, global model. Five centres (columns). Colour: macro F1 (\%), green = high, red = low.}
  \label{fig:S10}
\end{figure}

\end{document}